\documentclass[sigconf]{acmart}

\usepackage{balance}
\hypersetup{draft}
\usepackage{multirow} 
\usepackage{csquotes}

\AtBeginDocument{%
  }

\setcopyright{cc}
\setcctype{by}
\copyrightyear{2026}
\acmYear{2026}
\acmDOI{10.1145/YYYYYYY.YYYYYYY}
\acmConference[ACMSE 2026]{2026 ACM Southeast Conference}{April 23--25, 2026}{Troy, AL, USA}
\acmBooktitle{2026 ACM Southeast Conference (ACMSE 2026), April 23--25, 2026, Troy, AL, USA}
\acmISBN{979-8-4007-2062-8/2026/04}

\begin{document}
\pagestyle{empty}

\title{From Phase Grounding to Intelligent Surgical Narratives}

\author{Ethan Peterson}
\affiliation{%
  \institution{New Mexico Institute of Mining and Technology}
  \city{Socorro}
  \state{New Mexico}
  \country{USA}
}
\email{ethan.peterson@student.nmt.edu}

\author{Huixin Zhan}
\affiliation{%
  \institution{New Mexico Institute of Mining and Technology}
  \city{Socorro}
  \state{New Mexico}
  \country{USA}
}
\email{huixin.zhan@nmt.edu}


\begin{abstract}

Video surgery timelines are an important part of tool-assisted surgeries, as they allow surgeons to quickly focus on key parts of the procedure. Current methods involve the surgeon filling out a post-operation (OP) report, which is often vague, or manually annotating the surgical videos, which is highly time-consuming. Our proposed method sits between these two extremes: we aim to automatically create a surgical timeline and narrative directly from the surgical video.
To achieve this, we employ a CLIP-based multi-modal framework that aligns surgical video frames with textual gesture descriptions. Specifically, we use the CLIP visual encoder to extract representations from surgical video frames and the text encoder to embed the corresponding gesture sentences into a shared embedding space. 
We then fine-tune the model to improve the alignment between video gestures and textual tokens. Once trained, the model predicts gestures and phases for video frames, enabling the construction of a structured surgical timeline.
This approach leverages pretrained multi-modal representations to bridge visual gestures and textual narratives, reducing the need for manual video review and annotation by surgeons.

\end{abstract}



\maketitle

\section{Introduction}
In recent years, the surgical field has been moving toward minimally invasive surgeries, where the surgery is performed in the patient’s body cavity using small incisions to create a small wound through which a camera and tools can be inserted; this is usually done with robotic assistance \cite{LaparoscopicSURG}. These minimally invasive surgeries have been shown to lead to improvements in patient outcomes because they cause less damage to patients and result in faster recovery times \cite{LaparoscopicSURG, ma-2022}. Minimally invasive surgeries are also particularly beneficial for high-risk patients, where minimizing disruption to the body is critical.

As more surgeries are performed using minimally invasive approaches, the ability to precisely document how a surgery unfolds, both temporally and semantically, has taken on greater importance. This documentation not only serves as a record for medical professionals but is also critical for improving surgical practice and post-operative analysis. Traditional surgical timelines generally follow one of two paths. The surgeon can fill out a quick post-operative report that is little more than a few sentences describing the surgery \cite{Laccourreye2017OperativeReports}. While this approach is not ideal, it is the most commonly used option in practice due to surgeons’ time constraints. On the opposite end of the spectrum, in some medical schools, the surgeon can annotate the full surgical video to help students learn. While full annotation is valuable, it requires a significant amount of expert effort to complete accurately.

In response to the limitations of these two reporting methods, we propose a novel approach to enhance surgical documentation through the integration of machine learning techniques. Specifically, we intend to use Contrastive Language–Image Pre-Training (CLIP) \cite{radford2021learningtransferablevisualmodels}. CLIP generates a shared embedding space for a variety of text and images, allowing the model to ground images in language. By fine-tuning CLIP on a dataset of surgical gestures and phases, we aim to enable the model to associate surgical video clips with descriptive textual labels that capture the precise actions performed. This multi-modal embedding process allows for a rich, language-grounded representation of surgical activities.

\begin{figure}[H]
  \centering
  \includegraphics[width=\linewidth]{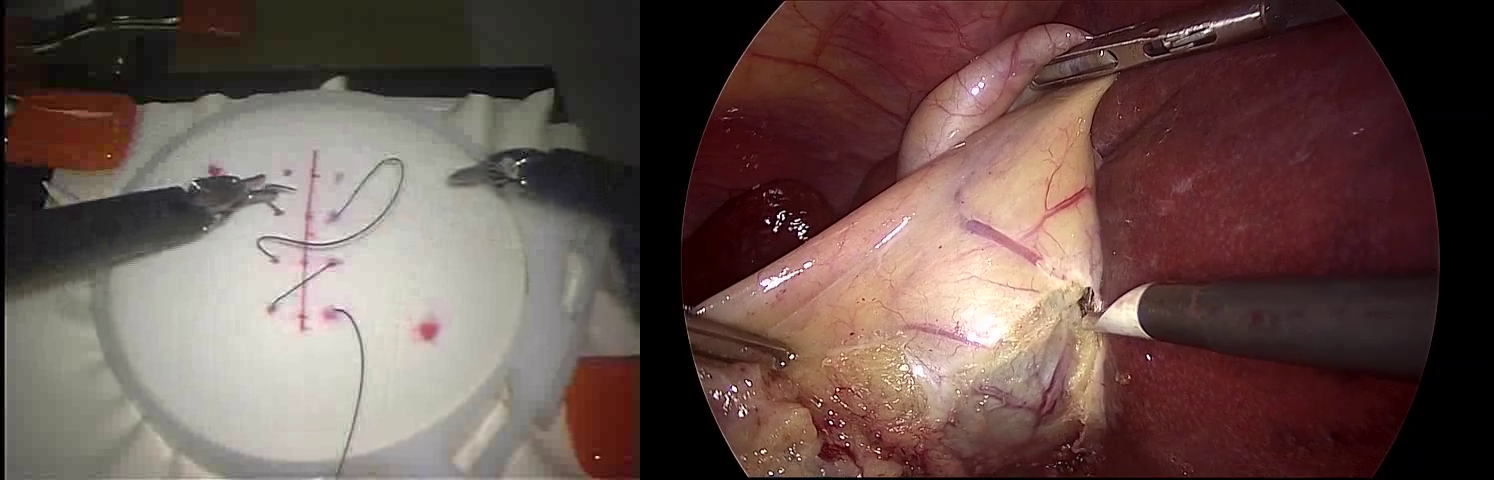}
  \caption{An Example Frame of the JIGSAWS Dataset \cite{gao2014jigsaws} (On the Right), and the Cholec80 Dataset \cite{twinanda2016endonetdeeparchitecturerecognition} (On the Left)}
  \label{fig:sidebyside} 
\end{figure}

As shown in Figure~\ref{fig:sidebyside}, although the JIGSAWS and Cholec80 datasets are both surgery-related, they differ substantially in appearance, with JIGSAWS featuring a sterile, white background and Cholec80 consisting of real-world surgical videos.

Our novelty lies in grounding surgical videos in language by using Contrastive Language Image Pre-Training \cite{radford2021learningtransferablevisualmodels}. We will do this by fine-tuning the CLIP model on surgical gestures and phases. By embedding surgical clips and textual labels into a shared semantic space, the model links visual patterns to interpretable language concepts such as \enquote{clipping and cutting} or \enquote{needle handling.} This grounding enables transparent, human-understandable summarization of surgical activities. While some previous research has used CLIP to create a shared embedding space for phase recognition, our novelty lies in using a multi-step process to align the shared embedding space. We propose to first fine-tune the CLIP model on surgical gestures using the JIGSAWS dataset. This will ground the model and allow it to recognize gestures. With the CLIP model aligned on gestures, we can then further fine-tune the model to detect the phase. This approach allows us to obtain a language-grounded representation of surgical videos.

\section{Related Works}

There has been substantial research on surgical phase recognition; however, most existing approaches follow one of two primary directions. The first direction involves recent multimodal methods that leverage richly annotated surgical datasets containing frame-level labels, kinematic data, and tool information. Rather than relying on a single annotation type, these approaches integrate multiple sources of information to enable a more comprehensive understanding of surgical scenes \cite{yuan2025hecvlhierarchicalvideolanguagepretraining, wang2025surgvidlmmultigrainedsurgicalvideo, li2024llavasurgmultimodalsurgicalassistant}. More recent models, such as VidLPRO \cite{honarmand2024vidlprounderlinevideounderlinelanguageunderlinepretrainingframework}, further incorporate audio signals to enhance surgical scene understanding. While the collection of such multimodal data can be challenging, the field is increasingly moving in this direction. One important application of surgical phase recognition models is their use in accurately predicting patient recovery and identifying critical moments during surgery where increased surgeon attention may be required \cite{li2025endendaisurgical}.

The second path taken by existing models is to rely solely on sequences of video frames and their corresponding labels to determine the surgical phase, using a variety of temporal modeling techniques. Early work in surgical phase recognition employed Hidden Markov Models (HMMs) \cite{PADOY2012632}, which enabled the modeling of temporal dependencies between surgical phases. Building upon these approaches, later methods introduced Convolutional Neural Networks (CNNs) to extract frame-level features, followed by a separate temporal model to infer the surgical phase from a sequence of frames \cite{yengera2018moresurgicalphaserecognition}. More recent work in this area has leveraged Transformer-based architectures, often combined with additional temporal modeling components, to determine surgical phases from sequences of video frames \cite{zhang2024friendstimemultiscaleaction, Loukas2019}.

All of the aforementioned methods have their respective advantages and limitations; however, one particularly promising area of research involves language-grounded models that construct a shared embedding space to align linguistic descriptions with surgical video or frame-level data. Despite its potential, relatively limited work has explored this direction. The model MML-SurgAdapt \cite{walimbe2025adaptationmultimodalrepresentationmodels} pursues a related objective; however, similar to other multimodal approaches, it does not rely exclusively on surgical video and frame data. To date, the only widely adopted language-grounded model in this domain is SurgCLIP \cite{perez2025surglavilargescalehierarchicaldataset}, which is supported by a newly introduced large-scale hierarchical dataset.

\section{Methodology}
Our proposed method involves several steps in the timeline creation. First, we will fine-tune the CLIP model to recognize gestures. Using this gesture-aware CLIP model, we then further fine-tune it to determine the surgical phase associated with each frame. Finally, we use that embedding along with a transformer to determine the phase of a sequence.

\subsection{Baseline Model}
To fine-tune the CLIP model on gestures, we first took a CLIP model, i.e. ViT-B/32, as the backbone. This is a model that uses the architecture of a vision transformer with a patch size of 32x32 ~\cite{dosovitskiy2021imageworth16x16words}. This backbone is well suited for our task, as the Vision Transformer variant of CLIP has been shown to perform strongly across a wide range of tasks \cite{radford2021learningtransferablevisualmodels}, and its Transformer-based design is particularly appropriate for identifying surgical gestures and phases.

The vision Transformer model operates by dividing an image into a sequence of patches, which in the case of ViT-B/32 are 32x32 pixels \cite{dosovitskiy2021imageworth16x16words}. Self-attention is then applied to model relationships among these patches, enabling the network to capture global image context. It has been shown that Transformer attention mechanisms can approximate the behavior of convolutional layers \cite{li2021visiontransformersperformconvolution}. As a result, Vision Transformers often outperform convolutional neural networks in many visual recognition tasks.

\begin{figure}[H]
  \centering
  \includegraphics[width=.99\linewidth]{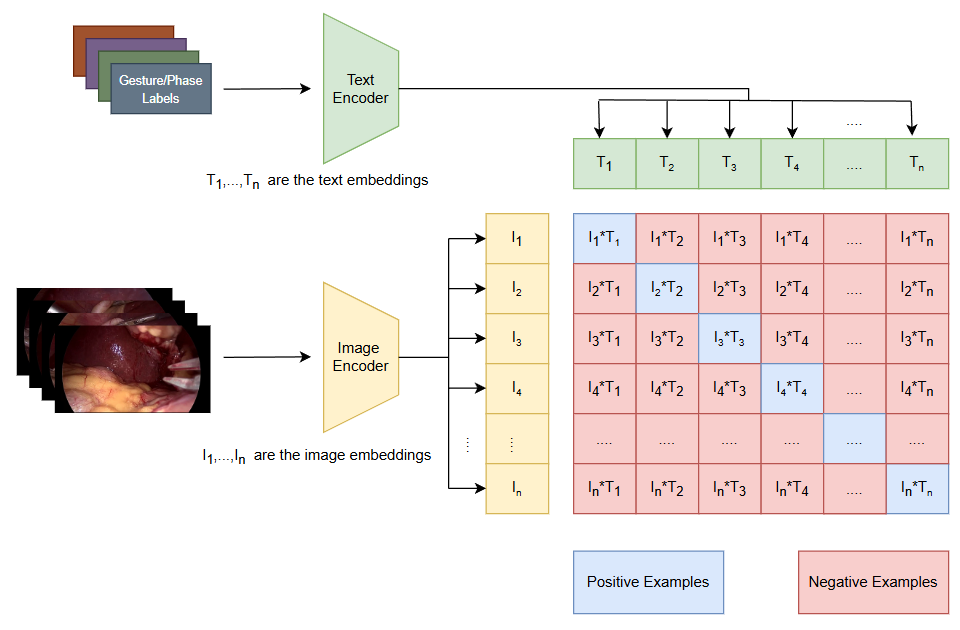}
  \caption{CLIP Framework Aligning Video Frames and Text in a Shared Embedding Space, Adapted from \cite{radford2021learningtransferablevisualmodels}}
  \label{fig:clip_aligning}
\end{figure}

The CLIP model was pre-trained by OpenAI on a large-scale dataset that contained 400 million image-text pairs obtained from the internet. This vast quantity of data allows the CLIP model to learn a wide range of visual features, enabling images to be grounded in language. This explicit relationship between text and image allows the model to generalize to unseen data extremely well. An illustration how CLIP aligns textual labels with the corresponding video frames is shown in Figure~\ref{fig:clip_aligning}.

\subsection{Datasets}
In our research on grounding surgical videos in language, we have selected two main surgical datasets. The first one is the JIGSAWS dataset \cite{gao2014jigsaws}. It consists of 15 surgical gestures. The gestures are part of one of three subsets, which contain videos of surgical tasks: suturing, needle passing, and knot tying. The Surgical tasks were performed by eight surgeons with differing levels of skill, and the videos were then hand annotated as one of the 15 gestures by an experienced surgeon. The Suturing task involves performing a simple suture, the knot tying task involves tying a single loop knot, and the needle passing task involves passing the needle between loops using the graspers \cite{gao2014jigsaws}. While these simplified tasks demonstrate several fundamental surgical gestures, and kinematic data are available, in this work we use only the gesture labels.

The gestures in the JIGSAWS dataset and the phases in the Cholec80 dataset are not very good descriptions. If we used the gesture or phase ID directly, it would lead to suboptimal understanding, as P1 or G1 is not highly descriptive. To address the issue, we created text banks for the JIGSAWS and Cholec80 datasets. These text banks contain a canonical description for gestures or phases. While using a single canonical description might work for simple tasks, we created an additional four paraphrases for each canonical description in order to better ground video frames in language. For example, the first gesture in the JIGSAWS dataset canonically is \enquote{Reaching for the needle with the right hand.} A paraphrase might be \enquote{The tool in the right hand reaches for the needle.} All of the canonical descriptions are in the Table ~\ref{tab:gesture_des}. While most of them are highly descriptive, there are a few overlapping gestures, such as G6 and G7, where the primary difference is which hand pulls the suture.

\begin{table}[!htbp]
\centering
\footnotesize
\caption{Surgical Gestures in the JIGSAWS Dataset \cite{gao2014jigsaws}, Mapped to Their Canonical Descriptions}
\label{tab:gestures}
\begin{tabular}{p{0.14\linewidth} | p{0.78\linewidth}}
\hline
Gesture & Gesture Description \\
\hline
G1  & Reaching for the needle with the right hand. \\[1mm]
G2  & Positioning the tip of the needle. \\[1mm]
G3  & Pushing needle through the tissue. \\[1mm]
G4  & Transferring needle from left to right. \\[1mm]
G5  & Moving to center of workspace with needle in grip. \\[1mm]
G6  & Pulling suture with the left hand. \\[1mm]
G7  & Pulling suture with the right hand. \\[1mm]
G8  & Orienting needle. \\[1mm]
G9  & Using right hand to help tighten a suture. \\[1mm]
G10 & Loosening more suture. \\[1mm]
G11 & Dropping the suture and moving to end points. \\[1mm]
G12 & Reaching for the needle with the left hand. \\[1mm]
G13 & Making a C loop around the right hand. \\[1mm]
G14 & Reaching for suture with right hand. \\[1mm]
G15 & Pulling suture with both hands. \\[1mm]
\hline
\end{tabular}
\label{tab:gesture_des}
\end{table}

The second dataset used in this research is the Cholec80 dataset \cite{twinanda2016endonetdeeparchitecturerecognition}, which contains videos of complete laparoscopic cholecystectomy procedures recorded at the University Hospital of Strasbourg, France. Laparoscopic cholecystectomy is a common surgical procedure performed on patients with gallbladder disease to remove the gallbladder. It is a minimally invasive, robot-assisted surgery that aims to reduce surgical trauma while achieving effective gallbladder removal. From the Cholec80 dataset, we used only the phase annotations in conjunction with the surgical videos in order to remain consistent with our use of the JIGSAWS dataset. Similar to the JIGSAWS dataset, we constructed a text bank consisting of one canonical description and four paraphrases for each phase label. In the case of the Cholec80 dataset, this resulted in a canonical descriptions, as shown in Table~\ref{tab:phase_des}.

\begin{table}[H]
\centering
\footnotesize
\caption{Surgical Phases of the Cholec80 Dataset \cite{twinanda2016endonetdeeparchitecturerecognition}, Mapped to Their Canonical Descriptions}
\label{tab:phases}
\begin{tabular}{p{0.14\linewidth} | p{0.78\linewidth}}
\hline
Phase & Phase Description \\
\hline
P1 & Positioning instruments inside the abdominal cavity to begin the procedure. \\[1mm]
P2 & Dissecting the Calot triangle to expose the cystic duct and artery. \\[1mm]
P3 & Applying clips and transecting the cystic duct and artery. \\[1mm]
P4 & Separating the gallbladder from the liver bed. \\[1mm]
P5 & Placing the gallbladder into a retrieval pouch for extraction. \\[1mm]
P6 & Cleaning the surgical site and coagulating bleeding vessels. \\[1mm]
P7 & Retracting the gallbladder to improve exposure of the operative field. \\[1mm]
\hline
\end{tabular}
\label{tab:phase_des}
\end{table}

\subsection{Loss Function}
\label{sec:less_func}
To ground the surgical videos in language, CLIP uses InfoNCE loss \cite{oord2019representationlearningcontrastivepredictive,radford2021learningtransferablevisualmodels}, which takes all of the images in the batch and uses the exact pairing as the positive example and the rest of the batch as negative examples. The use of positive and negative pairs allows alignment of the text and image encoders. In our case, where we fine-tune CLIP, the training procedure must be modified. When CLIP is trained from scratch, it leverages thousands of labels , that leads to good generalization. In our case, when we train the model on the JIGSAWS dataset, we only have 15 classes of images, and for the Cholec80 dataset, we only have 7 classes. To address this small quantity of classes, we adopt a multi-positive contrastive loss formulation, in which all image–text pairs belonging to the same class are treated as positive examples, while all remaining image–text pairs are treated as negative examples.

\subsection{Fine-Tuning the CLIP Model on Gestures}
To recognize individual frames, we sampled one frame every five frames from the JIGSAWS dataset. Since JIGSAWS videos are recorded at 30 frames per second (fps), this resulted in an effective frame rate of 6 fps. Prior to fine-tuning, we first verified whether the CLIP model could recognize surgical images at all. Although CLIP was trained on a large and diverse dataset, there were concerns that it might not generalize well to surgical gestures. To perform this verification, we trained a single-layer linear probe to classify surgical gestures.

After confirming that fine-tuning was a viable approach, we partitioned the JIGSAWS dataset into 60\% training, 10\% validation, and 30\% testing sets. The split was performed at the video level to prevent data leakage.

We selected these splits because the JIGSAWS dataset is highly imbalanced, and we aimed to retain a sufficiently large test set. To mitigate class imbalance during training, we applied up-sampling by matching all classes to the maximum class frequency through the creation of additional samples for underrepresented classes. It is important to note that while class imbalance can negatively impact performance, the primary objective at this stage is to ground surgical gesture videos in language. As such, moderate imbalance is not catastrophic for this phase, since the focus is on learning frame-to-language alignment rather than achieving optimal classification accuracy. 

\textbf{Settings:} To fine-tune the model, we took the \enquote{clip-vit-base-patch32} model, and froze all of its weights. We then unfroze the last three layers of the text and image encoders. This means that most of the model was frozen. We fine-tuned it for a total of 50 epochs, and we used an Adam optimizer with a learning rate of 5e-5. We had a batch size of 64. We used the modified InfoNCE loss function mentioned in Section ~\ref{sec:less_func}.

When training the linear probe to verify that the model can be adapted to surgical tasks we trained it for 200 epochs with a learning rate of 
5e-4, using the Adam optimizer and cross-entropy loss. The dataset was split into 60\% training, 10\% validation, and 30\% testing sets.

\subsection{Further Fine-Tuning the CLIP Model on Surgical Phases}
\label{sec:finetuningPhases}

At this stage, we moved from the JIGSAWS dataset to the Cholec80 dataset \cite{twinanda2016endonetdeeparchitecturerecognition}. This dataset contains 80 videos covering the complete laparoscopic cholecystectomy procedure. While this represents a substantial number of videos, and training on the full dataset was feasible, our focus on language grounding led us to use only a subset comprising 16.25\% of the dataset. Specifically, we selected 13 videos from Cholec80 and sampled every fifth frame, resulting in an effective frame rate of 5 fps.

Of the 13 surgical videos, 9 were used for training, 1 for validation, and 3 for testing. The surgical phase classes were heavily imbalanced, necessitating class balancing. Given the long duration of the surgical videos, down-sampling was chosen as the preferred balancing strategy. To perform down-sampling, we identified the class with the fewest examples and randomly sampled frames from the remaining classes until they matched this count. After balancing the training set, all frames were preprocessed and normalized according to the CLIP model’s input requirements.

\textbf{Settings:} To fine-tune the CLIP model, we took the model previously fine-tuned on the JIGSAWS dataset and similarly froze all of the weights. We then unfroze the last three layers of the text and image encoders. We fine-tuned it for 15 epochs with the Adam optimizer using a learning rate of 5e-5. We had a batch size of 32, and we used the loss function mentioned in Section ~\ref{sec:less_func}, which is a modified InfoNCE loss function. The reason for the small batch size is this set of training data is actually smaller than the JIGSAWS dataset due to the down-sampling. The small size is because some phases are short especially phases like gallbladder retraction.

For comparison, to check how our model that was fine-tuned on both JIGSAWS and Cholec80 was performing, we fine-tuned the base model on just the Cholec80 dataset for 15 epochs with the same settings mentioned above. Beyond that, we also fine-tuned the base model for 65 epochs on just the Cholec80 dataset to make the training time equivalent to both stages of fine-tuning.

\section{Experimental Results}
All of the experiments were performed using a single A100 NVIDIA 40GB GPU. We primarily evaluated performance using top 1 accuracy. For phase recognition, we extended that to top 5 accuracy. The experiments include training a linear probe, fine-tuning CLIP on gestures, and further fine-tuning on phases. The results are shown below.

\subsection{Results of Training the Linear Probe}

This result is expected for a single-layer linear probe and shows that the CLIP model is capable of learning gestures. The validation accuracy starts to level off around 200 epochs, yielding a total accuracy of 62.01\%, which demonstrates that CLIP can learn surgical gestures.

\subsection{Results of Fine-Tuning the CLIP Model on Gestures}

\begin{table}[!htbp]
\centering
\caption{Comparison Between Base Model and Model Fine-Tuned on JIGSAWS}
\begin{tabular}{|l|c|c|}
\hline
Metric     & Base CLIP Model & Fine-Tuned Model \\
\hline
Accuracy   & 3.05\%          & 59.17\% \\
Precision  & 19.29\%         & 65.29\% \\
Recall     & 3.05\%          & 59.17\% \\
F1 Score   & 4.23\%          & 61.10\% \\
\hline
\end{tabular}
\label{tab:JIGSAWS_comparison}
\end{table}

The result in Table~\ref{tab:JIGSAWS_comparison} is encouraging, as although an accuracy of 59.17\% is not perfect, it represents a substantial improvement over the base model's accuracy of 3.05\%. This improvement indicates that the fine-tuned model has successfully grounded surgical gestures in language. The fine-tuned model achieves performance similar to the linear probe despite having slightly less training data, which is desirable because we can train a linear probe on top of it if needed. We also computed the top 5 accuracy for each gesture to further verify that the model recognizes individual gestures. 

\begin{figure}[H]
  \centering
  \includegraphics[width=\linewidth]{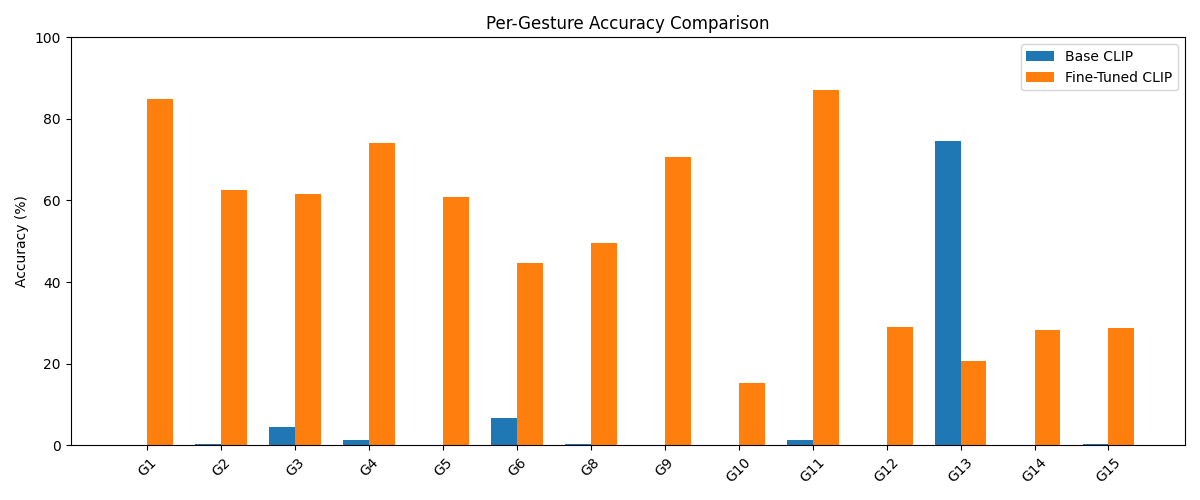}
  \caption{Comparison of Gesture Accuracy Between Base Model and Model Fine-Tuned on JIGSAWS}
  \label{fig:gesture_acc}
\end{figure}

As shown in Figure~\ref{fig:gesture_acc}, the fine-tuned model performs adequately for most gestures (G1–G9 and G11). Performance is lower for more complex gestures; however, given that this stage focuses on gesture–language grounding, the observed top-1 accuracy is acceptable, particularly since the fine-tuned model consistently outperforms the base model. A major factor contributing to the lower accuracy is class imbalance.

One notable outlier is gesture G13 (“Making a C loop around the right hand”), for which the base model outperforms the fine-tuned model. This behavior is likely due to the prevalence of “C-shaped” visual concepts in the original CLIP training data, enabling the base model to recognize this gesture effectively. After fine-tuning on the limited surgical dataset, the model appears to lose some of this general visual generalization, resulting in reduced performance on this particular gesture.

\subsection{Results of Further Fine-Tuning the CLIP Model on Surgical Phases}
To evaluate model performance, we compared accuracy using two metrics: top-1 accuracy, in which the model’s highest-confidence prediction must match the ground-truth label, and top-5 accuracy, in which the correct label must appear among the model’s five highest-confidence predictions.

We evaluated several models: the baseline CLIP model without any fine-tuning; the JIGSAWS fine-tuned model, which was fine-tuned exclusively on the JIGSAWS dataset; and the Cholec80-only model, which was obtained by fine-tuning the base CLIP model solely on the Cholec80 dataset. All of these models were compared against our final model, which was first fine-tuned on the JIGSAWS dataset and subsequently fine-tuned on the Cholec80 dataset.

\begin{table}[!ht]
\centering
\caption{Top-5 Accuracy per Gesture and Overall Accuracy for All Training Configurations. Best Result per Row is Bolded}
\resizebox{\columnwidth}{!}{%
\begin{tabular}{lcccc}
\toprule
\textbf{Metric} & \textbf{Baseline} & \textbf{JIGSAWS FT} & \textbf{Cholec80 Only FT} & \textbf{Cholec80 FT (Init: JIGSAWS)} \\
\midrule
\textbf{Overall Accuracy} & 0.3759 & 0.0395 & 0.2646 & \textbf{0.7035} \\
\midrule
Phase 1 Accuracy & 0.1518 & \textbf{0.9990} & 0.9920 & 0.6583 \\
Phase 2 Accuracy & 0.7090 & 0.0020 & 0.2123 & \textbf{0.7584} \\
Phase 3 Accuracy & 0.0312 & 0.0088 & 0.7943 & \textbf{0.5131} \\
Phase 4 Accuracy & 0.0078 & 0.0000 & 0.1458 & \textbf{0.7407} \\
Phase 5 Accuracy & 0.0065 & 0.0000 & 0.0562 & \textbf{0.6195} \\
Phase 6 Accuracy & 0.0000 & 0.0000 & 0.0609 & \textbf{0.4441} \\
Phase 7 Accuracy & 0.0000 & 0.0000 & 0.0000 & \textbf{0.6533} \\
\bottomrule
\end{tabular}%
}

\label{tab:top5}
\end{table}

For top-5 accuracy, Table~\ref{tab:top5} shows that our language-grounded model achieves the highest performance, with an accuracy of 70.35\%. This result indicates that the model reliably identifies the surgical phase associated with a given video frame, demonstrating that it is learning language-grounded representations of surgical frames.

\begin{table}[!ht]
\centering
\caption{Top-1 Accuracy per Gesture and Overall Accuracy for All Training Configurations. Best Result per Row is Bolded}
\resizebox{\columnwidth}{!}{%
\begin{tabular}{lcccc}
\toprule
\textbf{Metric} & \textbf{Baseline} & \textbf{JIGSAWS FT} & \textbf{Cholec80 Only FT} & \textbf{Cholec80 FT (Init: JIGSAWS)} \\
\midrule
\textbf{Overall Accuracy} & 0.0330 & 0.1223 & 0.1951 & \textbf{0.7025} \\
\midrule
Phase 1 Accuracy & 0.0040 & 0.7789 & \textbf{0.9628} & 0.6563 \\
Phase 2 Accuracy & 0.0280 & 0.1726 & 0.0287 & \textbf{0.7561} \\
Phase 3 Accuracy & 0.0053 & 0.0190 & 0.7192 & \textbf{0.5116} \\
Phase 4 Accuracy & 0.0032 & 0.0058 & 0.2480 & \textbf{0.7424} \\
Phase 5 Accuracy & 0.0314 & 0.0000 & 0.1816 & \textbf{0.6130} \\
Phase 6 Accuracy & 0.0058 & 0.0000 & 0.1702 & \textbf{0.4465} \\
Phase 7 Accuracy & 0.5061 & 0.0292 & 0.0000 & \textbf{0.6545} \\
\bottomrule
\end{tabular}%
}
\label{tab:top1}
\end{table}

\begin{figure}[!htbp]
  \centering
  \includegraphics[width=.72\linewidth]{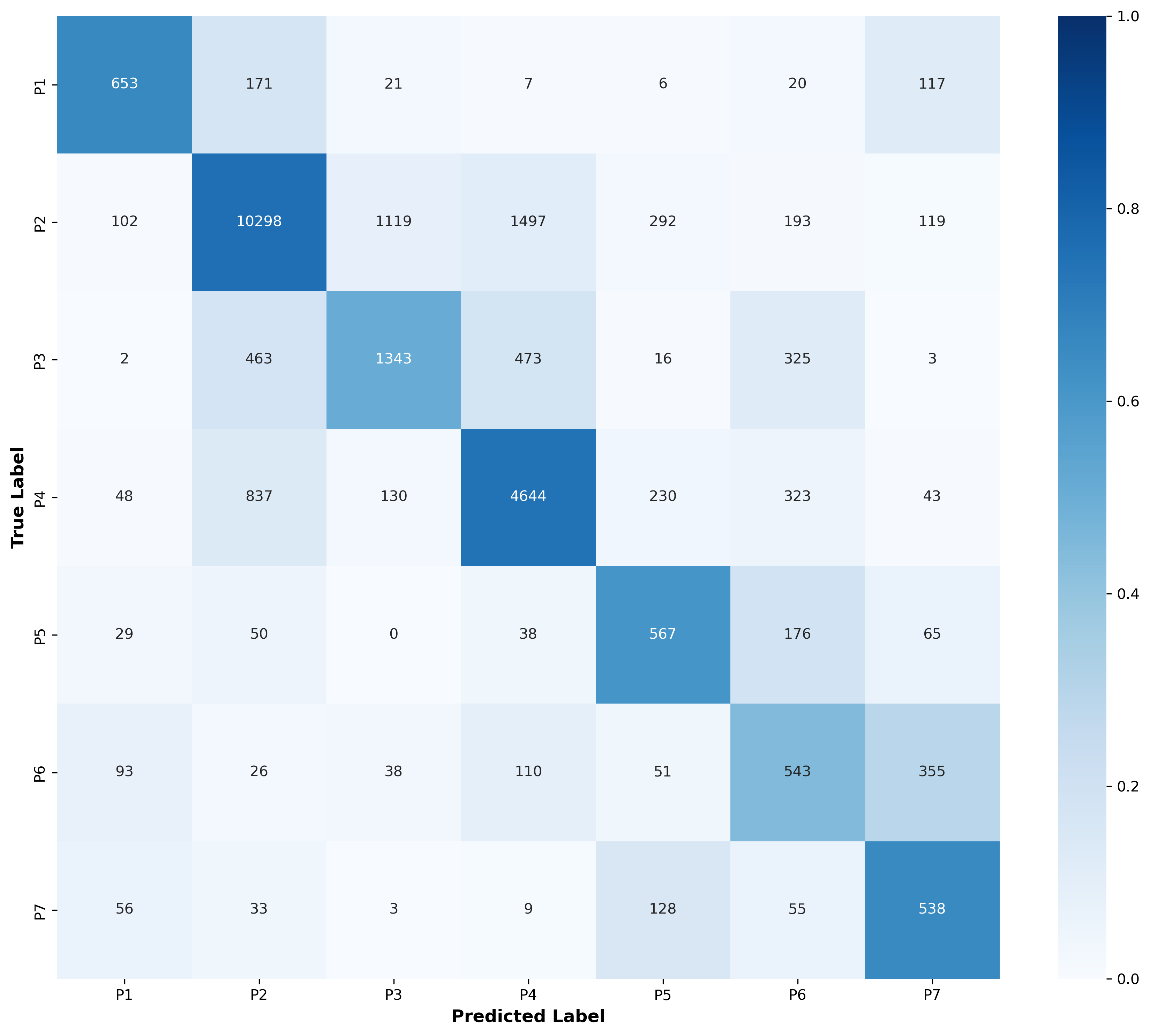}
  \caption{Confusion Matrix with the Colors Normalized}
  \label{fig:confusion_matrix}
\end{figure}

The results presented in Table~\ref{tab:top1} indicate that the model fine-tuned on both the JIGSAWS and Cholec80 datasets effectively grounds surgical video frames in language. In contrast, the model fine-tuned only on the Cholec80 dataset does not appear to learn the phase labels effectively. 
To better understand the model’s limitations, we refer to Figure~\ref{fig:confusion_matrix}, which presents the confusion matrix. The results indicate that the model struggles with Phase 3, where it confuses different cutting-related phases, and Phase 6, where it has difficulty distinguishing between cleaning the surgical site and retracting the gallbladder.

\begin{figure}[!htbp]
  \centering
  \includegraphics[width=\linewidth, height=0.1\textheight, keepaspectratio=false]{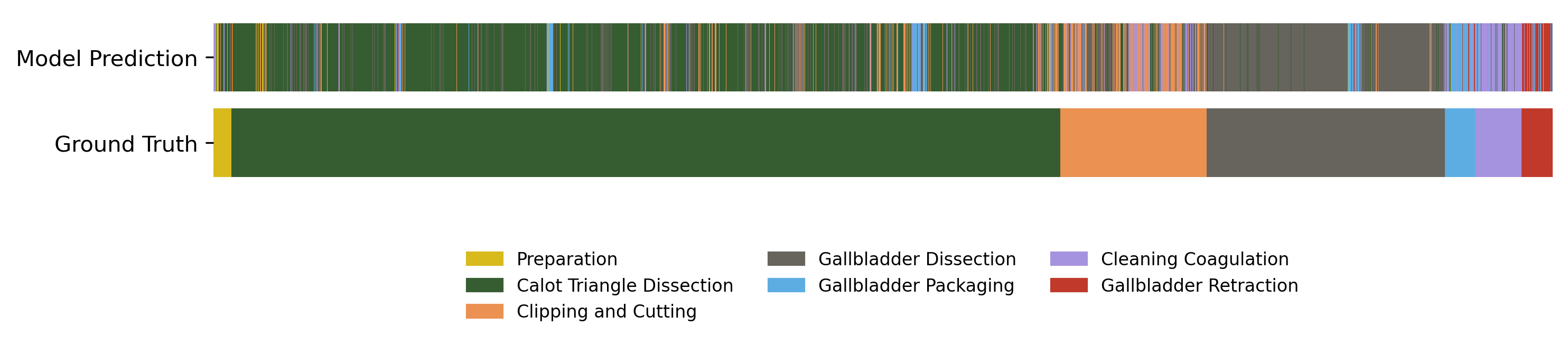}
  \caption{Model Prediction vs Ground Truth}
  \label{fig:phase_diagram}
\end{figure}

A more comprehensive way to evaluate the model’s overall performance is to examine the phase diagram, which compares the model’s predictions with the ground truth across an entire surgical video, as shown in Figure~\ref{fig:phase_diagram}.

\subsection{Results of Further Fine-Tuning the Base Model}

One concern was that the model fine-tuned on the JIGSAWS dataset might perform better simply due to the increased training time allocated to it. To assess this possibility, we trained the base CLIP model for 65 epochs on the Cholec80 dataset.

\begin{table}[!ht]
\centering
\caption{Combined Top-1 and Top-5 Accuracies for All Evaluated Cholec80 Configurations. Best Result per Row is Bolded}
\label{tab:combined_acc}
\resizebox{\columnwidth}{!}{
\begin{tabular}{lccc}
\toprule
\multicolumn{4}{c}{\textbf{Top-1 Accuracy}} \\
\midrule
\textbf{Metric} & \textbf{Ch80 FT (Init: JIGSAWS)} & \textbf{Ch80 Only FT} & \textbf{Ch80 Only FT (Full)} \\
\midrule
\textbf{Overall Accuracy} & \textbf{0.7025} & 0.1951 & 0.1411 \\
Phase 1 Accuracy & 0.6563 & \textbf{0.9628} & 0.6472 \\
Phase 2 Accuracy & \textbf{0.7561} & 0.0287 & 0.0141 \\
Phase 3 Accuracy & 0.5116 & 0.7192 & \textbf{0.9893} \\
Phase 4 Accuracy & \textbf{0.7424} & 0.2480 & 0.0000 \\
Phase 5 Accuracy & \textbf{0.6130} & 0.1816 & 0.2832 \\
Phase 6 Accuracy & \textbf{0.4465} & 0.1702 & 0.0304 \\
Phase 7 Accuracy & \textbf{0.6545} & 0.0000 & 0.0000 \\
\midrule
\multicolumn{4}{c}{\textbf{Top-5 Accuracy}} \\
\midrule
\textbf{Metric} & \textbf{Ch80 FT (Init: JIGSAWS)} & \textbf{Ch80 Only FT} & \textbf{Ch80 Only FT (Full)} \\
\midrule
\textbf{Overall Accuracy} & \textbf{0.7035} & 0.2646 & 0.4833 \\
Phase 1 Accuracy & 0.6583 & \textbf{0.9920} & 0.6623 \\
Phase 2 Accuracy & \textbf{0.7584} & 0.2123 & 0.8227 \\
Phase 3 Accuracy & 0.5131 & \textbf{0.7943} & 0.3322 \\
Phase 4 Accuracy & \textbf{0.7407} & 0.1458 & 0.0000 \\
Phase 5 Accuracy & \textbf{0.6195} & 0.0562 & 0.0508 \\
Phase 6 Accuracy & \textbf{0.4441} & 0.0609 & 0.0025 \\
Phase 7 Accuracy & \textbf{0.6533} & 0.0000 & 0.0000 \\
\bottomrule
\end{tabular}
}

\end{table}

When training the base model for 65 epochs, we observed that it did not effectively learn the gestures, as shown in Table~\ref{tab:combined_acc}. In contrast to the model that was further fine-tuned after initial training on the JIGSAWS dataset, the top-1 and top-5 accuracies for the base model remained notably different, indicating that the model was not learning the surgical phases effectively. These results suggest that language grounding using the JIGSAWS dataset leads to improved model performance.

\section{Limitations and Future Work}

In this work, we evaluated only partial fine-tuning of the CLIP model, where three layers of both the image and text encoders were unfrozen during training. Given that the CLIP ViT-B/32 architecture consists of 12 layers in the image encoder and 12 layers in the text encoder, this corresponds to fine-tuning approximately one quarter of the full model. As a result, the performance of the model under full fine-tuning remains unexplored. Future work will investigate fine-tuning the entire CLIP model and leveraging a larger portion of the Cholec80 dataset to further improve language grounding and phase recognition performance. Additionally, future work will extend the current frame-based approach to operate on sequences of frames, enabling more robust temporal modeling for surgical phase prediction.

\section{Conclusion}

In this work, we demonstrate that grounding surgical videos in language enables effective domain transfer in surgical video understanding. By fine-tuning CLIP in a staged manner, first at the level of surgical gestures and subsequently at the level of surgical phases, we encourage the model to progressively align visual representations with semantically meaningful linguistic concepts. Our experimental results show that this sequential fine-tuning strategy consistently outperforms direct fine-tuning on surgical phases alone. These findings suggest that intermediate language grounding at the gesture level provides a strong semantic foundation that facilitates improved generalization and transfer across surgical datasets and tasks.

\balance

\section*{Acknowledgments}
This research is supported by an Institutional Development Award (IDeA) from the National Institute of General Medical Sciences of the National Institutes of Health under grant number P20GM103451.

\bibliographystyle{ACM-Reference-Format}
\bibliography{sample-base}

\end{document}